\RequirePackage{fix-cm}
\documentclass[twocolumn]{svjour3}          % twocolumn
\smartqed  % flush right qed marks, e.g. at end of proof
\usepackage{graphicx}
\usepackage{wrapfig}
\usepackage{color}

\usepackage{float}
\usepackage{amsmath}
\usepackage{amssymb}
\usepackage[sort,numbers]{natbib}
\usepackage{diagbox}
\usepackage{algorithm}
\usepackage{algpseudocode}

\begin{document}
    \newpage

% For all papers calls, please submit 8-12 pages.
\begin{sloppypar}

\title{Edge Aware Learning for 3D Point Cloud}
% \subtitle{Segment Any Building}
\author{Lei Li}
\institute{lilei@di.ku.dk, University of Copenhagen}
\date{ }% The correct dates will be entered by the editor

\maketitle

\begin{abstract}
 This paper proposes an innovative approach to \textbf{H}ierarchical \textbf{E}dge \textbf{A}ware 3D Point Cloud Learning (HEA-Net) that seeks to address the challenges of noise in point cloud data, and improve object recognition and segmentation by focusing on edge features. In this study, we present an innovative edge-aware learning methodology, specifically designed to enhance point cloud classification and segmentation. Drawing inspiration from the human visual system, the concept of edge-awareness has been incorporated into this methodology, contributing to improved object recognition while simultaneously reducing computational time. Our research has led to the development of an advanced 3D point cloud learning framework that effectively manages object classification and segmentation tasks. A unique fusion of local and global network learning paradigms has been employed, enriched by edge-focused local and global embeddings, thereby significantly augmenting the model's interpretative prowess. Further, we have applied a hierarchical transformer architecture to boost point cloud processing efficiency, thus providing nuanced insights into structural understanding. Our approach demonstrates significant promise in managing noisy point cloud data and highlights the potential of edge-aware strategies in 3D point cloud learning. The proposed approach is shown to outperform existing techniques in object classification and segmentation tasks, as demonstrated by experiments on ModelNet40 and ShapeNet datasets.
\keywords{3D Point Cloud \and Edge Learning \and Classification \and Segmentation}
\end{abstract}

% polish and expand for one paragraph with more academic

% Note: If you are using BibTeX, please use the following code:
 
\begin{figure}
    \centering
    \includegraphics[width=\columnwidth]{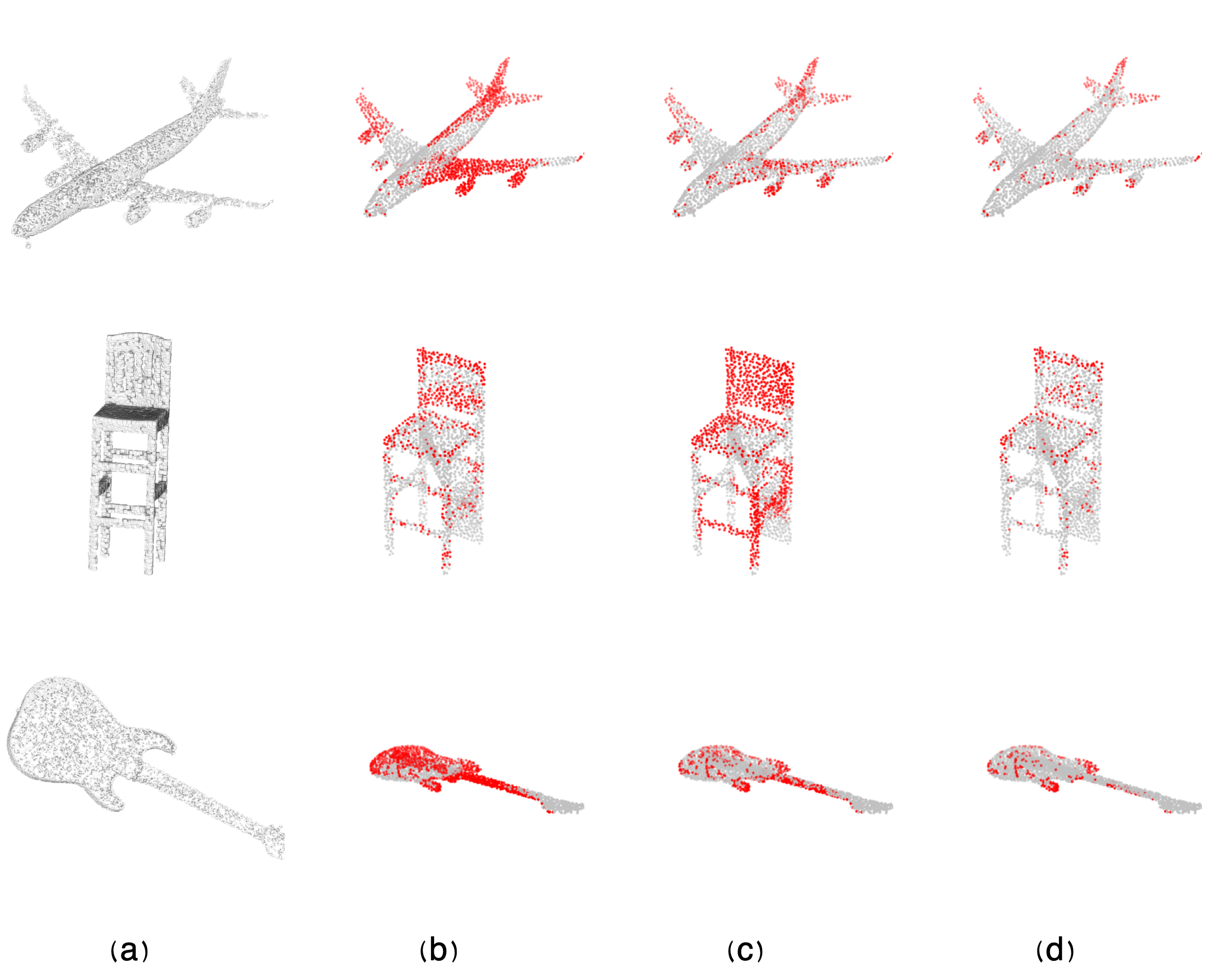}
    \caption{Our study applies edge-aware sampling to point cloud data at varying complexities—1024, 512, and 256 points—demonstrating consistent preservation of critical object geometry and topology information, irrespective of the reduction in point complexity. (a) Original point cloud; (b, c, d) edge-aware sampling.}
    \label{fig:sampling-step}
\end{figure}
\section{introduction}
Point clouds, a versatile data representation format, are central to numerous fields such as autonomous driving, augmented reality, and robotics. With the rise of advanced 3D sensing technologies, the acquisition of 3D data has become notably accessible, leading to the creation of an array of online data repositories, including ModelNet\cite{wu20153d}, ShapeNet\cite{chang2015shapenet}, and ScanNet\cite{dai2017scannet}. Amidst the variety of 3D shape representations, including voxel, mesh, and multi-view images, point clouds stand out as the preliminary data captured by LiDAR or depth cameras, like Kinect. The robust utility of point cloud data, underpinned by the critical task of representative point sampling, is evidenced in its broad applications, from scene reconstruction and autonomous driving navigation to virtual reality. As such, the development of efficient point cloud learning algorithms remains a significant focus in contemporary scientific and technological research.

Within the realm of 3D data processing, point cloud sampling methodologies, such as random sampling (RS), grid sampling, uniform sampling, farthest point sampling (FPS)\cite{eldar1997farthest}, and geometric sampling, are foundational. RS, while boasting efficiency, tends to overlook sparser regions, and FPS, despite its wide coverage, is hindered by latency issues during parallel operations. The application of grid sampling, through its use of regular grids, lacks the ability to accurately control the number of sampled points. Uniform sampling, renowned for its robust nature, evenly distributes the points chosen across the cloud. Geometric sampling strategies that focus on local geometric properties, like shape curvature, along with techniques like Inverse Density Importance Sampling (IDIS) that favor points with lower cumulative distances to neighbors, call for high-density point clouds for maximum efficacy.

Yet, in spite of the advancements brought about by these traditional methodologies, the preservation of geometric information during the sampling process, particularly for complex objects with nuanced topology structures and irregular surface morphology, remains a daunting challenge. This obstacle has led to the emergence of cutting-edge learn-to-sample methods that concurrently optimize both the sampling process and specific downstream tasks. Such innovations have made substantial contributions to domains such as point cloud classification and reconstruction. Nonetheless, the challenge of maintaining geometric integrity within complex objects is still a largely unexplored area in academic research. In light of the vast diversity of object topological structures, defining a consistent category-agnostic edge at the semantic level poses a significant non-trivial challenge. Current datasets typically encompass only singular or a limited range of known object categories. Additionally, topological methods, which are ordinarily category-agnostic, concentrate predominantly on the geometric and topological attributes of the shape, such as its connectivity, topology, length, direction, and width.

With the advancement of deep learning techniques, several neural network-based sampling methods, such as S-Net~\cite{dovrat2019learning}, SampleNet\cite{lang2020samplenet}, and DA-Net\cite{lin2021net}, have emerged. These approaches leverage multi-layer perceptrons (MLPs) to produce resampled point cloud sets of prescribed sizes, with MOPS-Net offering an innovation of generating a sampling transformation matrix. Despite their novelty, these methods lean toward a generative approach, bypassing direct point selection. Meanwhile, research has grown around creating neural network-based local feature aggregation operators for point clouds. However, these techniques, while reducing point numbers during latent feature learning, don't strictly qualify as sampling methods as they lack real spatial points during processing. Furthermore, the complex task of skeleton extraction adds another layer of complexity due to the inherent diversity of object topological structures and sensitivity to surface noise.

By leveraging deep neural networks to learn object edges, a prior probability for sampling each point, mindful of edge-awareness, can be established. To adapt edge-aware sampling for subsequent tasks, we synergistically optimize the posterior sampling probability for each point. This process is achieved in an end-to-end manner, utilizing the transformative capabilities of transformer neural network architectures.

Actually, the performance of machine learning models is often constrained by the presence of noisy points within the cloud, thereby calling for the evolution of techniques capable of robustly managing such data. The concept of edge-awareness is inspired by the human visual system that profoundly relies on edge or part information for object recognition. The integration of such edge-aware strategies into point cloud learning can augment recognition capacities and concurrently decrease computational time. In this study, we introduce an edge-aware learning methodology that significantly bolsters point cloud classification and segmentation, contributing in three keypoints:

\begin{itemize}
    \item We propose an innovative, edge-aware 3D point cloud learning framework handling object classification and segmentation.
    \item The fusion of local and global network learning paradigms, further supplemented by corresponding edge-focused local and global embeddings that enrich the model's interpretive abilities.
    \item The application of a hierarchical, transformer architecture that enhances point cloud processing efficiency, offering sophisticated insight into structural understanding.
\end{itemize}

\section{Related Work}

\paragraph{Point Cloud Sampling.}
The paramountcy of point cloud sampling in processing high-resolution dense point clouds is undeniable, and this recognition has precipitated the development of an array of innovative methods. Some techniques, for instance, harness K-means clustering to highlight representative points, concurrently removing redundancy. Others combine the power of clustering with a coarse-to-fine methodology to enhance point cloud processing efficiency. Intrinsic point cloud algorithms, providing a guarantee of density, further contribute to this field, facilitating uniform and feature-sensitive resampling. However, these strategies tend to concentrate on subset identification based on geometric or topological criteria, often overlooking the downstream task implications during sampling.

In recent years, the advent of learn-to-sample approaches has heralded a new era in this domain. These methodologies introduce learning-based ~\cite{zhang2020fact, wu2019fase,lang2020samplenet, li2022mask}, task-oriented sampling strategies~\cite{thomas2019kpconv, zhang2023attention} explicitly designed to cater to downstream tasks like point cloud classification, retrieval, and reconstruction. Some approaches incorporate a learnable sampling and joint-training strategy, mitigating overfitting risks for specific tasks. Others innovatively embed a critical point layer, transferring points with the most functional features to subsequent layers to enable adaptive fusion sampling. Moreover, these learning-based techniques often outperform the traditional non-learning-based sampling methods, such as the well-known FPS, emphasizing the continued evolution in this field.

\paragraph{Point Cloud Representation Learning.}
Deep learning methodologies have become integral to point cloud analysis, exhibiting pervasive influence across a spectrum of areas such as point cloud classification/segmentation ~\cite{qi2017pointnet, qi2017pointnet++,chang2015shapenet,lang2020samplenet,yan2020pointasnl,muzahid2020curvenet,wang2019dynamic,xu2018spidercnn}, object detection/tracking \cite{engel2021point}, point cloud autoencoders~\cite{thomas2019kpconv}, generation, completion, regression\cite{oehmcke:22,revenga2022above,oehmcke2022deep}, and registration. The inherent complexity associated with the unstructured nature of point clouds, where the points are not positioned on a regular grid and exhibit varying degrees of mutual independence and inconsistent distances to neighboring points, presents significant challenges to the application of deep learning techniques.

An innovative breakthrough in this regard has been the advent of PointNet~\cite{qi2017pointnet}, a trailblazing strategy that applied deep neural networks to point sets for the first time. This framework converted the input into high dimensional space on a point-wise basis and subsequently employed a symmetric function to aggregate all point features, achieving a permutation-invariant global feature representation. This method led to the widespread adoption of the Multi-Layer Perceptron (MLP) block within many point cloud networks to learn pointwise representations. Then, PointNet++ \cite{qi2017pointnet++}, refined this approach further by considering local information, thereby enhancing the representational capacity and efficiency of the network. 
The Dynamic Graph CNN (DGCNN) ~\cite{wang2019dynamic}, proposed EdgeConv blocks to dynamically update the neighborhood information based on dynamic graphs, leading to superior point cloud analysis performance. In a similar vein, KPConv~\cite{thomas2019kpconv} proposed point-wise convolution operators for point cloud feature learning.

Alongside these convolution-based approaches, attention-based methods have also gained traction. For instance, Point Cloud Transformer (PCT) \cite{engel2021point, wang2021pst} pioneered the use of self-attention layers in place of encoder layers within the PointNet framework. Another notable attention-based network, the Lightn \cite{wang2022lightn}, uses offset attention blocks to capture contextual information with light-weight transformer. These recent advancements demonstrate the potential of attention-based methodologies in capturing both local and long-range contextual information in the point cloud domain.

\paragraph{Edge Learning.}
The incorporation of edge or skeleton learning has emerged as a fundamental aspect for capturing crucial properties such as geometry, topology~\cite{zhang2022lr}, and symmetry~\cite{10024907,zhou4425635multi,zhang2023attention,li2022buildseg} in objects, providing a concise and intuitive representation. This representation has demonstrated its efficacy across diverse computer vision tasks, encompassing shape recognition, reconstruction, segmentation, and point cloud completion. For instance, deep learning techniques have been proposed to leverage edge-awareness in generating mesh reconstructions of object surfaces from single RGB images. In the realm of image processing, the detection of edges typically relies on well-established techniques such as the Canny edge detector, which involves a multi-stage algorithm encompassing Gaussian filtering, intensity gradient computation, gradient magnitude thresholding, double thresholding, and edge suppression. These techniques enable the identification and extraction of salient edges while effectively suppressing weaker and disconnected ones, thereby enhancing the edge-aware analysis and interpretation capabilities.

\section{Methods}
We introduce our \textbf{H}ierarchical \textbf{E}dge \textbf{A}ware 3D Point Cloud Learning (HEA-Net) framework, which incorporates global and local learning for capturing edge information effectively. The framework uses deep tier learning, which processes the point cloud at different tiers to learn hierarchical features, similar to the workings of a human visual system. A detailed discussion of the architecture, layers, and algorithms involved is provided.

\begin{figure*}[h]
    \centering
    \includegraphics[width=\textwidth]{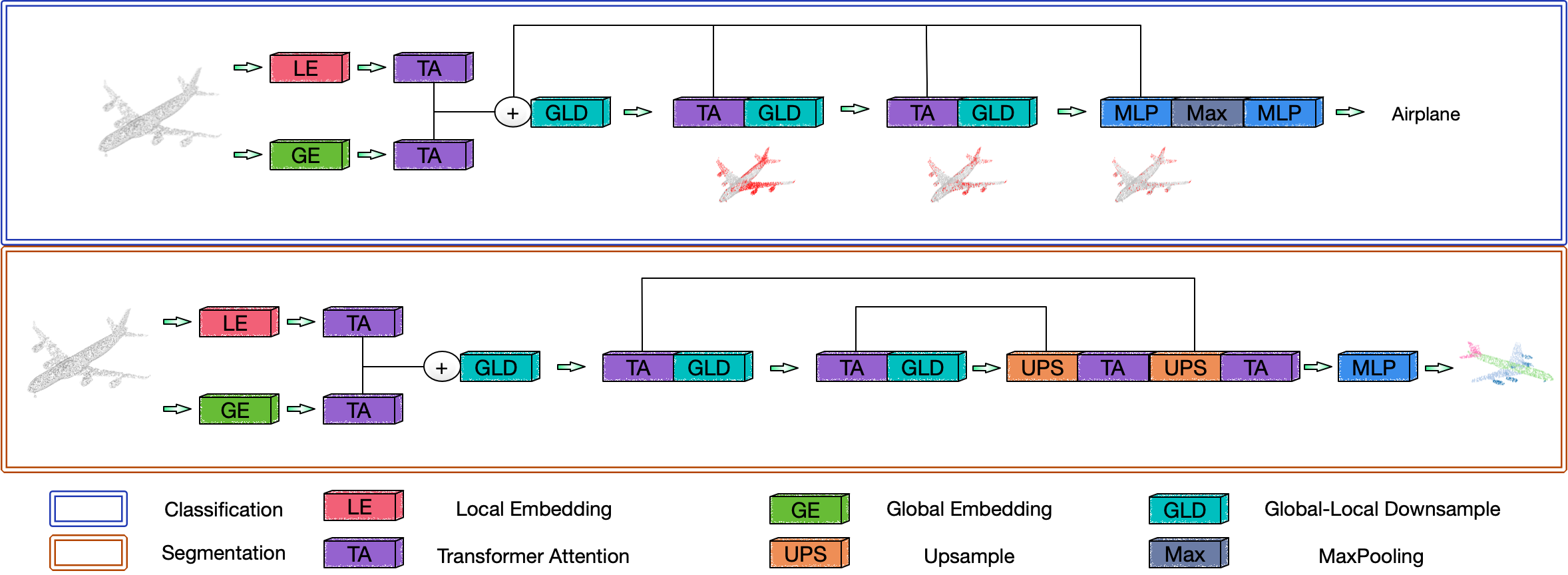}
    \caption{The Hierarchical Edge-Aware network (HEA-Net) architectures employ distinctive strategies for classification(above) and segmentation(below) tasks. For classification and segmentation, the architecture combines local and global embeddings to effectively learn edge features, followed by an innovative global-local downsampling process that reduces a point cloud's complexity from N to M points. On the other hand, in the context of segmentation, an upsample layer is introduced to augment the point cloud's complexity back from M to N points. Integral to the HEA network's process is the use of transformer-based attention mechanisms which enable the architecture to focus on and learn critical sampling features.}
    \label{fig:pipeline}
\end{figure*}
\subsection{Overview of Framework.}
The figure \ref{fig:pipeline} illustrates our proposed Hierarchical Edge-Aware Network (HEA-net), which is specifically designed to adeptly execute classification and segmentation tasks for point clouds. A key aspect to note is the consistent application of the downsampling process across both classification and segmentation tasks. Subsequent to the application of local and global embeddings, the resultant data is fused and then fed into a multi-level network, specifically the Global-Local Downsampling (GLD) module. Evidently, from the classification tasks, it is apparent that the multi-layer downsampling accentuates the contours and edges of the object (in this case, an airplane). This approach efficiently discards superfluous information, thereby enhancing the pipeline's accuracy and speed, making the HEA-net an effective tool for processing and interpreting point cloud data.

Within the scope of the classification task, the point cloud's maximal feature is ultimately classified via the max pooling module. Conversely, in the context of segmentation tasks, a sequence of downsampling followed by upsampling is executed prior to conducting the segmentation. It is worth noting that two skip connections have been incorporated within the entire network structure. These skip connections significantly enhance the network's ability to learn hierarchical features. Subsequently, we will delve into a detailed exposition of the embedding, downsampling, and transformer attention modules, further elucidating their pivotal roles in the overall architecture of the Hierarchical Edge-Aware Network.

\subsection{Embedding learning.}
Algorithm 1 succinctly delineates the process of obtaining local and global points through the CenterNeighbor and CenterDiff operations, where neighbor selection is conducted using the K-Nearest Neighbor (KNN) method. The number of neighbor points to be selected is pre-determined by assigning a value to K. Notably, the choice of K can influence the overall performance and speed of the process. Theoretically, there exists a critical value for K beyond which performance improvements plateau \cite{wu2023attention}.

Upon the selection of neighbor points, local features are sampled using these points, while global features are extracted by randomly selecting distant points that are not included within the neighborhood. It's crucial to note that this extraction process differs from the Farthest Point Sampling (FPS) method. Following the embedding of these points, multi-level feature learning is then applied, thus facilitating a more nuanced understanding of the data at various scales.

% \begin{algorithm}
% \caption{CenterNeighbor and CenterDiff}
% \begin{algorithmic}
%     \Procedure{CenterNeighbor}{$pcd \in \mathbb{R}^{B \times C \times N}, K$}
%         \State Call \Call{SelectNeighbors}{$pcd, K, "neighbor"$} to get $neighbors \in \mathbb{R}^{B \times C \times N \times K}$
%         \State Repeat $pcd$ K times across a new dimension to get $pcd_{repeated} \in \mathbb{R}^{B \times C \times N \times K}$
%         \State Concatenate $pcd_{repeated}$ and $neighbors$ to get $output \in \mathbb{R}^{B \times 2C \times N \times K}$
%     \Return $output$
%     \EndProcedure
%     \Procedure{CenterDiff}{$pcd \in \mathbb{R}^{B \times C \times N}, K$}
%         \State Call \Call{SelectNeighbors}{$pcd, K, "diff"$} to get $diff \in \mathbb{R}^{B \times C \times N \times K}$
%         \State Repeat $pcd$ K times across a new dimension to get $pcd_{repeated} \in \mathbb{R}^{B \times C \times N \times K}$
%         \State Concatenate $pcd_{repeated}$ and $diff$ to get $output \in \mathbb{R}^{B \times 2C \times N \times K}$
%     \Return $output$
%     \EndProcedure
%     % \label{alg:Embedding}
% \end{algorithmic}
% \end{algorithm}

\begin{algorithm}
\caption{CenterNeighbor and CenterDiff Procedures}
\begin{algorithmic}
    \Procedure{CenterNeighbor}{$pcd \in \mathbb{R}^{B \times C \times N}, K$}
        \State $neighbors \gets$ \Call{SelectNeighbors}{$pcd, K, "neighbor"$} 
        \State $pcd_{repeated} \gets$ Repeat $pcd$ K times across a new dimension
        \State $output \gets$ Concatenate $pcd_{repeated}$ and $neighbors$ along channel dimension
    \Return $output$
    \EndProcedure
    \Procedure{CenterDiff}{$pcd \in \mathbb{R}^{B \times C \times N}, K$}
        \State $diff \gets$ \Call{SelectNeighbors}{$pcd, K, "diff"$} 
        \State $pcd_{repeated} \gets$ Repeat $pcd$ K times across a new dimension
        \State $output \gets$ Concatenate $pcd_{repeated}$ and $diff$ along channel dimension
    \Return $output$
    \EndProcedure
\end{algorithmic}
\end{algorithm}

\subsection{Global and Local Learning for the Edge.}

\begin{figure}[h]
    \centering
    \includegraphics[width=\columnwidth]{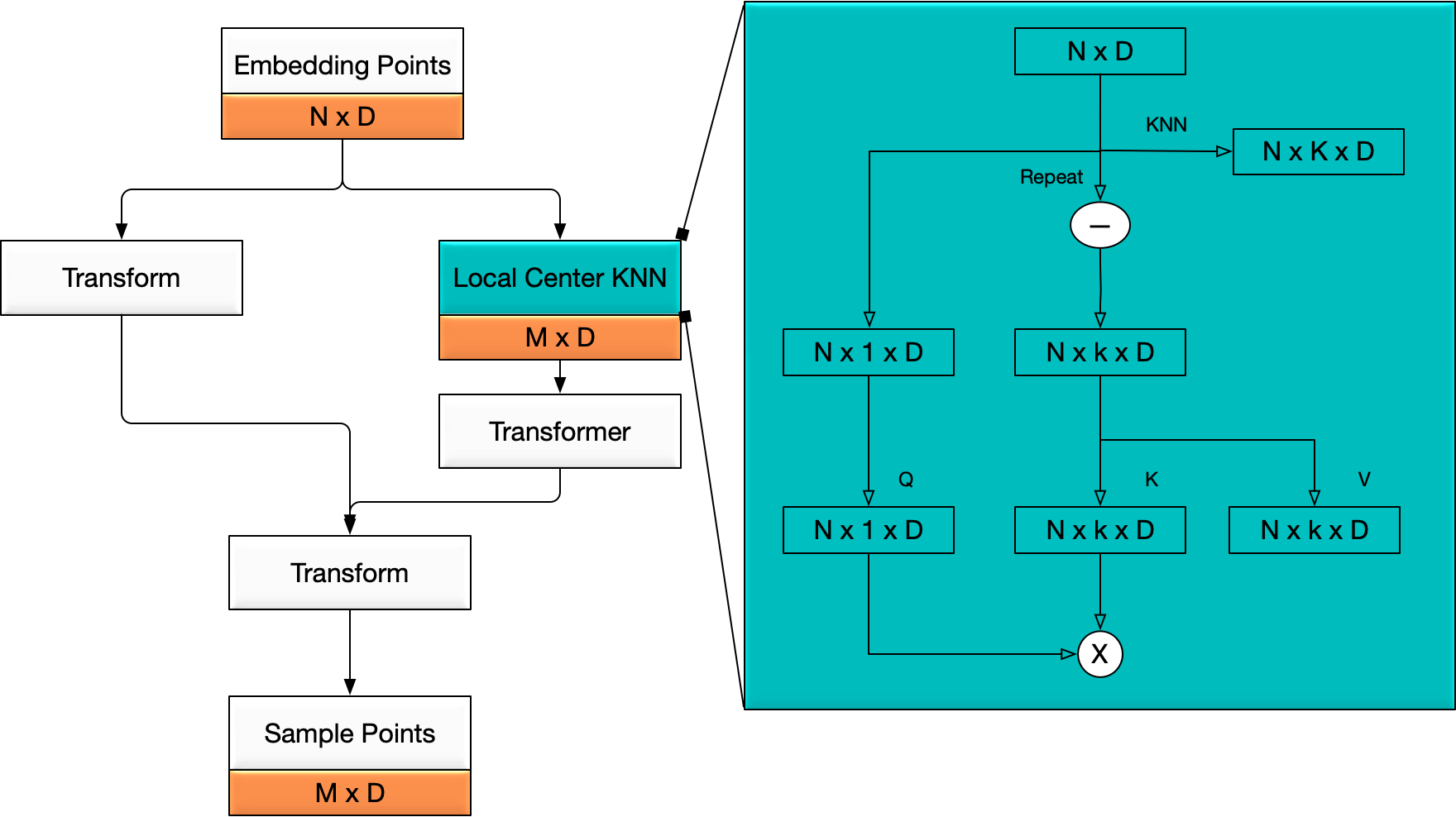}
    \caption{Hierarchical downsampling combined with global feature and local feature for the edge.}
    \label{fig:downsample}
\end{figure}
\begin{algorithm}
\caption{Hierarchical Global-Local DownSample Procedures}

\begin{algorithmic}
    \Procedure{GlobalDownSample}{$x \in \mathbb{R}^{B \times C \times N}$}
        \State Compute $q, k, v \gets$ Transformer operations on $x$
        \State Compute energy $E \gets q @ k$
        \State Compute attention $A \gets$ softmax($E$)
        \State Compute selection $S \gets$ sum $A$ across last dimension
        \State Select top $M$ indices $idx$ from $S$
        \State Select scores by gathering $A$ across index $idx$
        \State Compute output $out \gets$ scores $@$ $v$
    \EndProcedure
    \Statex
    \Procedure{LocalDownSample}{$x \in \mathbb{R}^{B \times C \times N}$}
        \State Compute neighbors of $x$ and $q, k, v \gets$ Transformer operations
        \State Compute energy $E \gets q @ k$
        \State Compute attention $A \gets$ softmax($E$)
        \State Compute selection $S \gets$ std. deviation of $A$
        \State Select top $M$ indices $idx$ from $S$
        \State Select scores by gathering $A$ across index $idx$
        \State Compute output $out \gets$ scores $@$ $v$
    \EndProcedure
    \Statex
    \Procedure{GlobalLocalDownSample}{$x \in \mathbb{R}^{B \times C \times N}$}
        \State Compute $global_out \gets$ \Call{GlobalDownSample}{$x$}
        \State Compute $local_out \gets$ \Call{LocalDownSample}{$x$}
        \State Compute output $out \gets global_out + local_out$
    \EndProcedure
\end{algorithmic}
\end{algorithm}

Algorithm 2 provides a detailed overview of the Hierarchical Global-Local Downsample algorithm. Q, K, and V parameters are derived utilizing a transformer-based approach, followed by a direct implementation of global correspondence learning. Conversely, the LocalDownsample process acquires neighbor points through the K-Nearest Neighbor (KNN) method, subsequently employing attention mechanisms to discern the relationships between the neighbor points and the feature points.

As depicted in the corresponding Figure \ref{fig:downsample}, point cloud information from the multi-scale fused embedding is integrated via local and global attention operations. Post-fusion, the dimension of the output point cloud stands at N * D. This output is then subjected to a downsampling procedure, yielding M*D data. So, combined with Embedding layers, we can use hierarchical global and local feature to downsampling the point cloud. The global and local correlation maps, denoted as $M^g$ and $M^l$, are combined into a single $N \times 2N$ correlation matrix $M$:

\begin{equation}
M = \begin{bmatrix}
- & m_1^{g\top} & - & - & m_1^{l\top} & - \\
- & m_2^{g\top} & - & - & m_2^{l\top} & - \\
 & \vdots & & & \vdots & \\
- & m_N^{g\top} & - & - & m_N^{l\top} & -
\end{bmatrix}
\end{equation}

In the combined correlation matrix $M$, let $m_{i j}^g$ and $m_{i j}^l$ represent the values at the $i$-th row and $j$-th column in the $M^g$ and $M^l$ partitions of $M$ respectively. If point $i$ is an edge point, its corresponding $m_i^g$ will have a larger standard deviation, indicating that point $j$ is also likely to be an edge point due to the large $m_{ij}$.

Instead of computing row-wise standard deviations as previously done, we propose to compute column-wise sums. Let $u_j = \sum_i m_{ij}$, where $m_{ij}$ are the elements of $M^g$. We sample points that have a higher value of $u_j$. The underlying rationale is that a point which contributes more to the self-attention correlation map as a respresentive feature.

\begin{figure}
    \centering
    \includegraphics[width=\columnwidth]{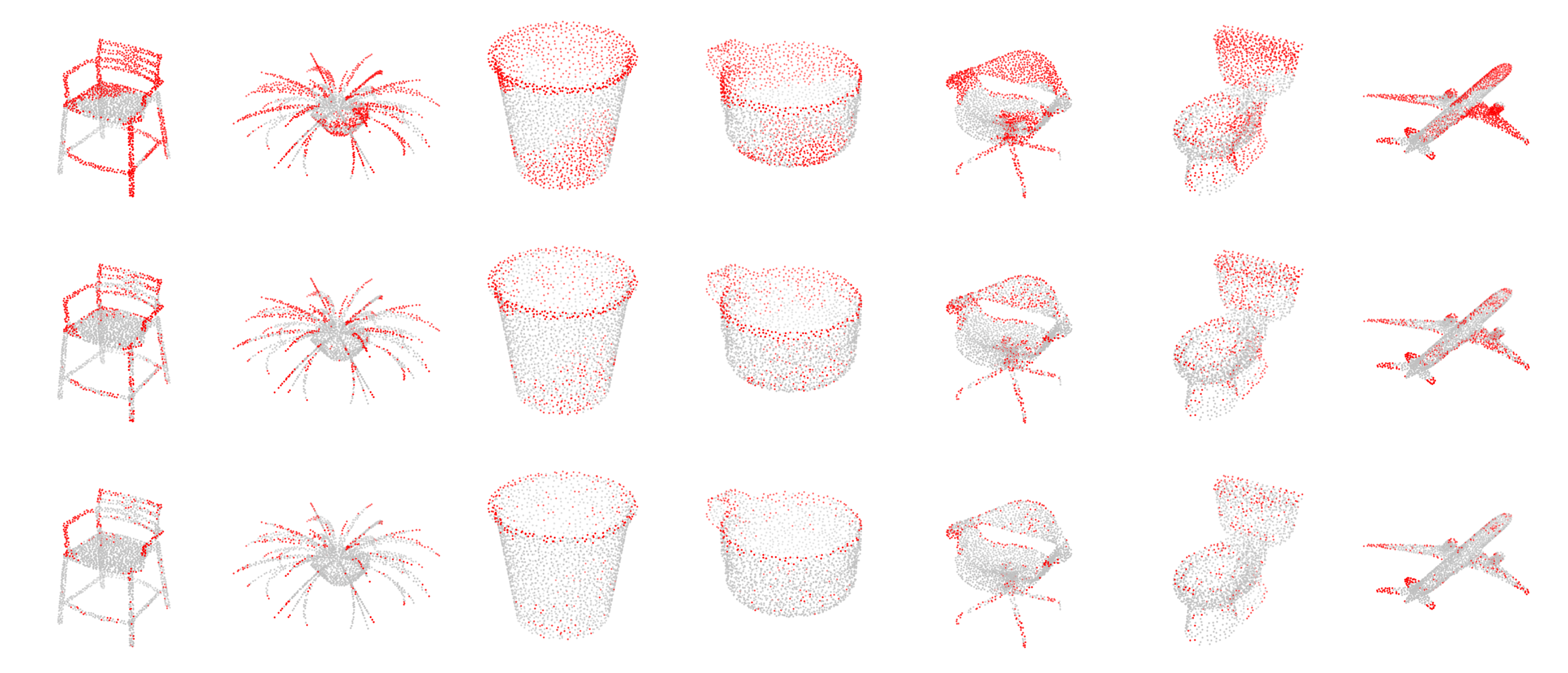}
    \caption{Hierarchical deep downsampling.The visual representation from top to bottom depicts the evolution of point cloud information following three stages of downsampling. The elements depicted in red symbolize the point cloud post-sampling. }
    \label{fig:deep-sampling}
\end{figure}

To create $M^l$, we use the (latent) features of the center point $p_i$ as the query input and the feature difference between the neighbor point and the center point $(p_{ij} - p_i)$ as the key input. As per the original Transformer model, we employ the square root of the feature dimension count ($\sqrt{d}$) as a scaling factor. The final normalized correlation map $m_l^i$ is thus given as follows:

As demonstrated in Figure \ref{fig:deep-sampling}, the feature information of the object post three hierarchical downsamplings and the final point cloud backcalculation is distinctly evident. Upon completion of the final sampling process, the M-dimensional information manifests as an effective representation of the object's information, highlighting the efficacy of the Hierarchical Global-Local Downsample algorithm in preserving essential data attributes. As we traverse from left to right, we observe the varying feature information expressed by different objects, underlining the capacity of this methodology to capture and illustrate a wide array of object-specific characteristics in an efficient and concise manner.

\subsection{Transformer Attention.}

The algorithm 3 begins with the definition of the Transformer Decoder, which calculates the self-attention of the input, computes multi-head attention of the self-attention and memory, and applies a feed-forward neural network. After that, the Transformer Decoder is applied within the Transformer Attention. The input is rearranged, transformed through the decoder, rearranged again, and then normalized twice before returning the output.

\begin{algorithm}
\caption{Transformer Attention}
\begin{algorithmic}
    \Require Input tensor $x \in \mathbb{R}^{B \times C \times N}$
    \Ensure Output tensor $y \in \mathbb{R}^{B \times C \times N}$
    
    \State Initialize Transformer Decoder with the necessary parameters \\
    \Comment{Step 1: Apply Transformer Decoder on rearranged input}
    \State Rearrange $x$ to have dimensions 'N B C'
    \State Apply Transformer Decoder on rearranged $x$ to get the transformed tensor
    \State Rearrange the transformed tensor back to  'B C N' \\
    \Comment{Step 2: Normalization and Finalization}
    \State(BN1) on the rearranged tensor
    \State(BN2) on the output of the previous step
    
    \Return Result from BN2
\end{algorithmic}
\end{algorithm}

\subsection{Loss function.}

In the field of 3D point cloud analysis, the CrossEntropyLoss function is frequently used to quantify the discrepancy between predicted class probabilities and the actual labels. This loss function is vital for training models involved in semantic segmentation or classification tasks on 3D point cloud data. For a single point within the point cloud, the CrossEntropyLoss is defined as:

\begin{equation}
L(y, \hat{y}) = - \sum_{c=1}^C y_c \cdot \log(\hat{y}_c)
\end{equation}

Here, $L(y, \hat{y})$ represents the CrossEntropyLoss between the true labels $y$ and the predicted labels $\hat{y}$. The notation $y_c$ stands for the true label of class $c$, and $\hat{y}_c$ denotes the predicted probability of class $c$. The logarithmic term intensifies the penalty for incorrect predictions, aiding the model to enhance its prediction accuracy.

\section{Experiments}
\subsection{Evaluation Metric.}
For Classification task, we use Accuracy to present the performance of the methods.
\begin{equation}
Acc = \frac{1}{N}\sum_{i=1}^{N} 1(y_i = \hat{y}_i)
\end{equation}
In this equation, $N$ denotes the total number of points, $y_i$ represents the actual class of the $i$-th point, and $\hat{y}_i$ is the predicted class of the $i$-th point. The function $1(.)$ is the indicator function, which equals to $1$ when $y_i = \hat{y}_i$ (i.e., when the prediction is correct), and $0$ otherwise.

In the realm of segmentation tasks, the metrics of Category Mean Intersection over Union (Cat.mIoU) and Instance Mean Intersection over Union (Ins.mIoU) are employed to evaluate the efficacy of our proposed methods.

Cat.mIoU is utilized to assess the performance of semantic segmentation models. It quantifies the intersection-over-union (IoU) for each category or class and computes the average over all categories. The Cat.mIoU is mathematically represented as:

\begin{equation}
Cat.mIoU = \frac{1}{C} \sum_{i=1}^{C} \frac{TP_i}{TP_i + FP_i + FN_i}
\end{equation}

In this equation, $C$ denotes the total number of categories or classes. $TP_i$, $FP_i$, and $FN_i$ represent the count of true positives, false positives, and false negatives for the category $i$, respectively.

IMIoU, on the other hand, is a metric employed for instance segmentation tasks. It measures the intersection-over-union (IoU) between predicted and ground truth instances, and calculates the mean across all instances. The Ins.mIoU is mathematically depicted as:

\begin{equation}
Ins.mIoU = \frac{1}{N} \sum_{j=1}^{N} \frac{TP_j}{TP_j + FP_j + FN_j}
\end{equation}

Here, $N$ corresponds to the total number of instances. $TP_j$, $FP_j$, and $FN_j$ signify the count of true positives, false positives, and false negatives for instance $j$, respectively. These metrics provide an encompassing overview of the model's performance across different aspects of the segmentation task.

\subsection{Data and setting}
To validate our method, we carry out a series of experiments using ModelNet40 and ShapeNet datasets. During the training phase, we employed the \text{AdamW} optimizer with an initial learning rate of $1 \times 10^{-4}$ and utilized a cosine annealing schedule to gradually decay the learning rate to $1 \times 10^{-8}$ over the course of 400 epochs, while maintaining a batch size of 16. Additionally, we incorporated a weight decay hyperparameter of $1 \times 10^{-5}$ to regulate the network weights, and introduced dropout with a probability of 0.5 in the last two fully connected layers. 
These specific choices of hyperparameters and regularization techniques were carefully selected to optimize the training process, ensure model convergence, and enhance the overall performance and generalization capability of the network.

\subsection{Results}

\paragraph{quantitative analysis.} A thorough quantitative analysis was conducted to evaluate the performance of our proposed methodology in both classification and segmentation tasks, utilizing the ModelNet40 and ShapeNet Part datasets respectively in Table~\ref{tab:ModelnetClass} and Table~\ref{tab:seg}. Despite operating with a reduced number of sample points, our method consistently exhibited superior performance, as substantiated by the empirical results. In essence, our approach demonstrated an exceptional capability to succinctly capture the intricate patterns inherent in the sampled point cloud outlines. This attests to its effectiveness as a tool for point cloud analysis, further emphasizing the significance of our contribution to the field.

\begin{table}[ht]
\centering
\caption{Classification performance on ModelNet40.}
\begin{tabular}{|c|c|}
\hline 
Method & Overall Accuracy \\
\hline 
PointNet \cite{qi2017pointnet} & 89.2 \% \\
\hline 
PointNet++ \cite{qi2017pointnet++} & 91.9 \% \\
\hline \cite{xu2018spidercnn} & 92.4 \% \\
\hline 
DGCNN \cite{wang2019dynamic} & 92.9 \% \\
\hline 
PointCNN \cite{li2018pointcnn} & 92.2 \% \\
\hline 
PointConv \cite{wu2019pointconv} & 92.5 \% \\
\hline 
PVCNN \cite{liu2019point} & 92.4 \% \\
\hline 
KPConv \cite{thomas2019kpconv} & 92.9 \% \\
\hline 
PointASNL \cite{yan2020pointasnl} & 93.2 \% \\
\hline 
PT$^1$ \cite{engel2021point} & 92.8 \% \\
\hline 
PT$^2$ \cite{zhao2021point} & 93.7 \% \\
\hline 
PRA-Net \cite{cheng2021net} & 93.7 \% \\
\hline 
PAConv \cite{xu2021paconv} & 93.6 \% \\
\hline 
CurveNet \cite{muzahid2020curvenet} & 93.8 \% \\
\hline 
DeltaConv \cite{wiersma2021deltaconv} & 93.8 \% \\
\hline 
% APES (local-based) & 93.47? \% \\
% \hline 
% APES (global-based) & 93.8 \% \\
% APES & 93.47\% \\
% LSA & 89.1\% \\
\hline
HEA-Net & 93.8 \\
% Edge-fuse-embed & \\
\hline
\end{tabular}
\label{tab:ModelnetClass}
\end{table}

\begin{table}[ht]
\centering
\caption{Segmentation results on ShapeNet Part.}
\begin{tabular}{lcc}
\hline Method & Cat. mIoU & Ins. mIoU \\
\hline PointNet \cite{qi2017pointnet} & $80.4 \%$ & $83.7 \%$ \\
PointNet++ \cite{qi2017pointnet++}& $81.9 \%$ & $85.1 \%$ \\
SpiderCNN \cite{xu2018spidercnn} & $82.4 \%$ & $85.3 \%$ \\
DGCNN \cite{wang2019dynamic} & $82.3 \%$ & $85.2 \%$ \\
SPLATNet \cite{su2018splatnet} & $83.7 \%$ & $85.4 \%$ \\
PointCNN \cite{li2018pointcnn} & $84.6 \%$ & $86.1 \%$ \\
PointConv \cite{wu2019pointconv} & $82.8 \%$ & $85.7 \%$ \\
KPConv \cite{thomas2019kpconv} & $85.0 \%$ & $86.2 \%$ \\
PT ${ }^1$ \cite{engel2021point}  & - & $85.9 \%$ \\
PT ${ }^2$ \cite{zhao2021point} & $83.7 \%$ & $8 6 . 6 \%$ \\
% PCT [12] & - & 86.4%86.4 \% \\
PRA-Net \cite{cheng2021net} & $83.7 \%$ & $86.3 \%$ \\
PAConv \cite{xu2021paconv} & $84.6 \%$ & $86.1 \%$ \\
CurveNet \cite{muzahid2020curvenet} & - & $86.6 \%$ \\
StratifiedTransformer \cite{lai2022stratified} & $85.1 \%$ & $8 6 . 6 \%$ \\
% \hline APES (local-based) & % \% & 84.69%84.69 \% \\
% APES (global-based) & % \% & 85.04%85.04 \% \\
% \hline APES (local-based) & 83.1%83.1 \% & 85.6%85.6 \% \\
% APES (global-based) & 83.7%83.7 \% & 85.8%85.8 \% \\
\hline
\hline
% Edge-fuse & % \% & 85.06% 85.06\% \\
% Edge-fuse-embed & % \% & % \% \\
HEA-Net & $ 83.9\%$ & $ 85.9\%$  \\

% 85.05
%  0.8515
% 85.85
\end{tabular}
\label{tab:seg}

\end{table}

\begin{figure*}[ht]
    \centering
    \includegraphics[width=\textwidth]{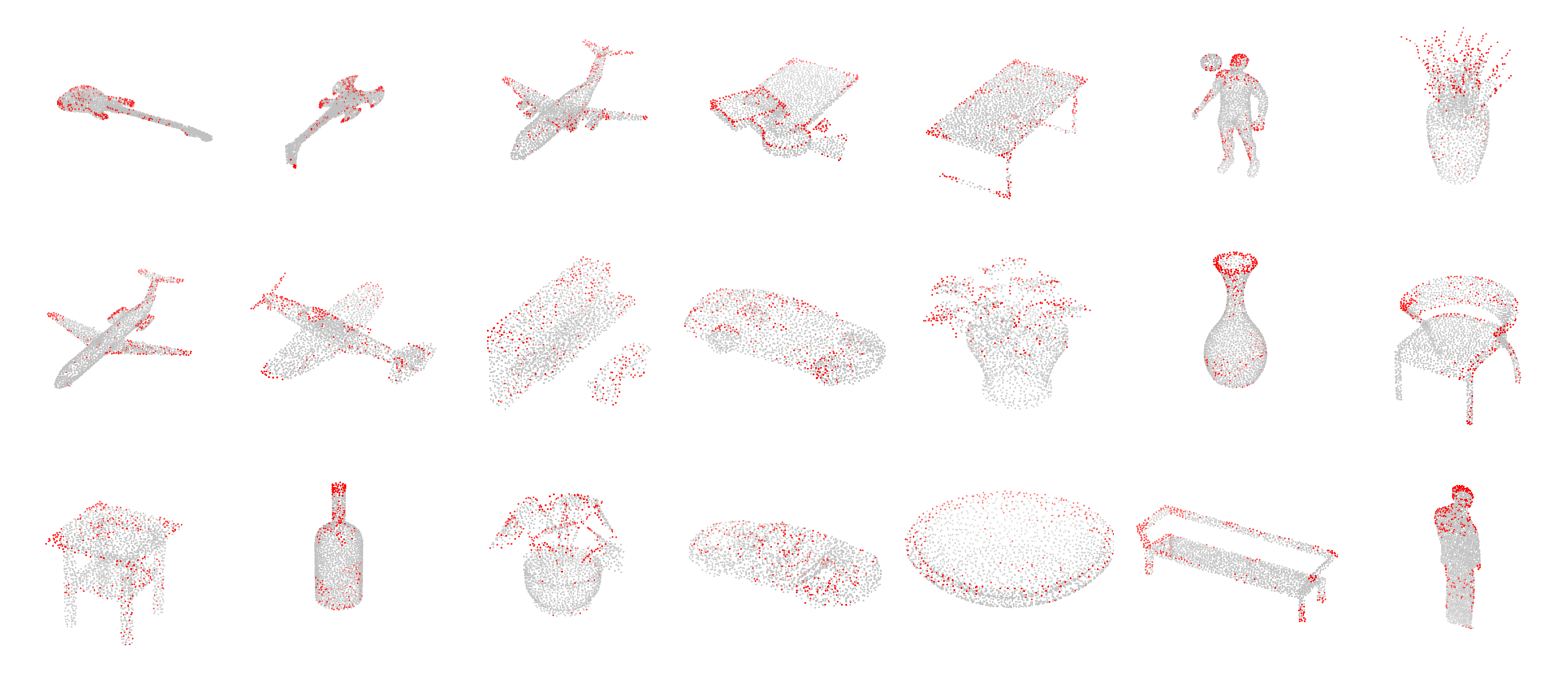}
    \caption{Visualization of varied shape sampling results are generated by HEA-Net with all shapes being drawn from the test set.}
    \label{fig:classification_predict}
\end{figure*}

\begin{figure*}[!ht]
    \centering
    \includegraphics[width=\textwidth]{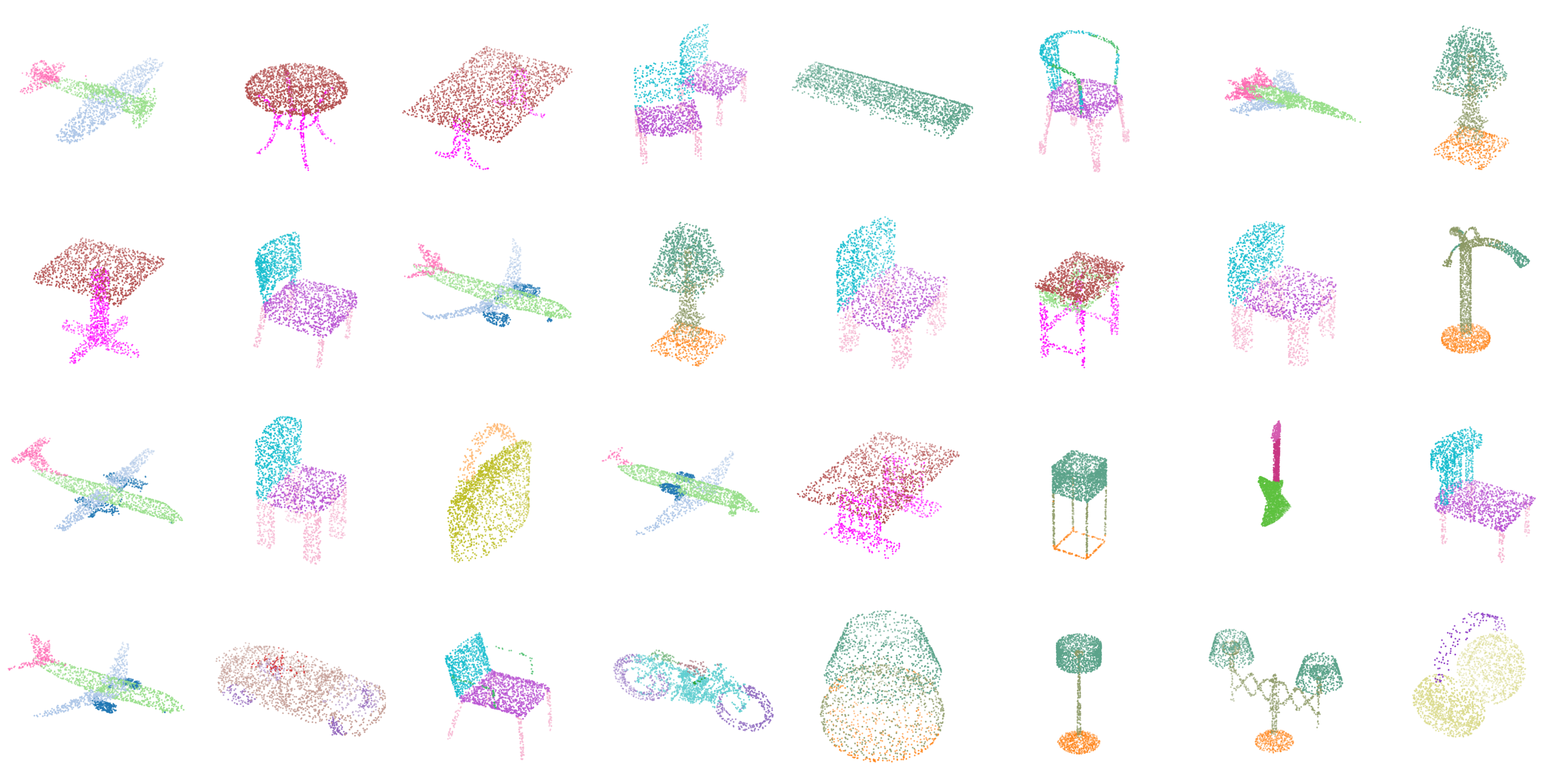}
    \caption{We provide a visualization of segmentation results as the point clouds of different shapes undergo downsampling. All shapes are from test set.}
    \label{fig:seg_predict}
\end{figure*}

\paragraph{Qualitative analysis.} Complementary to our quantitative findings in In Figure~\ref{fig:classification_predict} and Figure~\ref{fig:seg_predict}, we showcase a set of qualitative results, as illustrated in the accompanying figure. Through this visual representation, we aim to provide a comprehensive depiction of the performance characteristics of our proposed methodology. These results serve as valuable evidence to substantiate the efficacy of our approach in practical scenarios, offering a tangible and intuitive understanding of its capabilities. By presenting these qualitative findings, we enhance the comprehensibility and transparency of our research, enabling readers to gain deeper insights into the practical implications and potential applications of our proposed methodology.

\subsection{Ablation Study}

In order to conduct a rigorous comparative analysis, our proposed sampling technique is systematically evaluated alongside established methodologies, including Random Sampling (RS), Farthest Point Sampling (FPS), and recent learning-based approaches such as S-Net, SampleNet, and LightTN. This evaluation follows a standardized evaluation framework employed in prior research studies, as visually illustrated in Figure 8. The primary objective of this study is to assess the classification performance on the ModelNet40 dataset, utilizing the PointNet architecture as the baseline network. By systematically evaluating various sampling approaches across multiple sampling sizes, a comprehensive assessment of their relative performance and efficacy is conducted, contributing to the advancement of the field and providing valuable insights for future research endeavors.

% \begin{table*}[ht]
% \centering
% \caption{Our method is assessed against other prevalent sampling techniques, utilizing diverse sampling sizes within the ModelNet40 classification benchmark for a comprehensive comparison.}
% \resizebox{\textwidth}{!}{%
% \begin{tabular}{|c|c|c|c|c|c|c|c|c|c|c|c|c|}
% \hline$M$ & Voxel & $\mathrm{RS}$ & FPS \cite{eldar1997farthest} & S-NET \cite{dovrat2019learning} & PST-NET \cite{wang2021pst} & SampleNet \cite{lang2020samplenet} & MOPS-Net \cite{qian2020mops} & DA-Net \cite{lin2021net} & LighTN \cite{wang2022lightn} & APES \cite{wu2023attention} & HEA-Net \\
% \hline 512 & 73.82 & 87.52 & 88.34 & 87.80 & 87.94 & 88.16 & 86.67 & 89.01 & 89.91  & 90.81 & 93.75 \\
% \hline 256 & 73.50 & 77.09 & 83.64 & 82.38 & 83.15 & 84.27 & 86.63 & 86.24 & 88.21  & 90.40 & 92.26 \\
% \hline 128 & 68.15 & 56.44 & 70.34 & 77.53 & 80.11 & 80.75 & 86.06 & 85.67 & 86.26  & 89.77 & 90.24 \\
% % \hline 64 & 58.31 & 31.69 & 46.42 & 70.45 & 76.06 & 79.86 & 85.25 & 85.55 & 86.51 & 88.68 & 89.57 \\
% % \hline 32 & 20.02 & 16.35 & 26.58 & 60.70 & 63.92 & 77.31 & 84.28 & 85.11 & 86.18 & 86.49 & 88.56 \\
% \hline
% \end{tabular}
% }
% \label{tab:your_label}
% \end{table*}

\section{Discussion}

In our current research focus on indoor object analysis, we have applied our edge-aware learning method. However, extending this approach to outdoor objects presents unique challenges due to the complex and diverse nature of outdoor scenes. Outdoor environments encompass a wide range of factors, including varying lighting conditions, occlusions from natural elements, and intricate background structures, which complicate edge analysis. Effectively capturing hierarchical edge information becomes crucial for accurate outdoor object analysis. Nonetheless, addressing these challenges and extending edge-aware learning to outdoor contexts offers promising opportunities to enhance computer vision systems in real-world applications such as autonomous driving, surveillance, and environmental monitoring.

Additionally, striking a balance between achieving high accuracy and managing computational complexity is a key consideration. Future work involves expanding the applicability of our method to a broader range of applications, including outdoor scenarios, and exploring potential enhancements to our proposed framework. This entails investigating novel strategies to address the unique characteristics and challenges of outdoor scenes, as well as optimizing the trade-off between accuracy and computational efficiency. Such advancements will contribute to the advancement of edge-aware learning techniques and their broader adoption in diverse real-world scenarios.

\begin{table}[h]
\centering
\caption{Our method is assessed against other prevalent sampling techniques, utilizing diverse sampling sizes within the ModelNet40 classification benchmark for a comprehensive comparison.}
\begin{tabular}{|c|c|c|c|}
\hline
Method & $M=512$ & $M=256$ & $M=128$ \\
\hline
Voxel & 73.82 & 73.50 & 68.15 \\
\hline
RS & 87.52 & 77.09 & 56.44 \\
\hline
FPS \cite{eldar1997farthest} & 88.34 & 83.64 & 70.34 \\
\hline
S-NET \cite{dovrat2019learning} & 87.80 & 82.38 & 77.53 \\
\hline
PST-NET \cite{wang2021pst} & 87.94 & 83.15 & 80.11 \\
\hline
SampleNet \cite{lang2020samplenet} & 88.16 & 84.27 & 80.75 \\
\hline
MOPS-Net \cite{qian2020mops} & 86.67 & 86.63 & 86.06 \\
\hline
DA-Net \cite{lin2021net} & 89.01 & 86.24 & 85.67 \\
\hline
LighTN \cite{wang2022lightn} & 89.91 & 88.21 & 86.26 \\
\hline
APES \cite{wu2023attention} & 90.81 & 90.40 & 89.77 \\
\hline
HEA-Net & 93.75 & 92.26 & 90.24 \\
\hline
\end{tabular}
\label{tab:sample_ablation}
\end{table}

\section{Conclusion}
The human visual system relies extensively on edge and part information for object recognition, serving as inspiration for edge-awareness in computer vision. In this paper, we propose a novel framework for learning from 3D point cloud data, specifically addressing object classification and segmentation tasks. Our approach integrates the strengths of both local and global network learning paradigms, complemented by edge-focused local and global embeddings that enhance the model's interpretive capacities. Furthermore, we leverage a hierarchical, transformer-based architecture to improve the efficiency of point cloud processing and enable deeper structural comprehension. Through extensive experiments, we validate the efficacy of our proposed framework, demonstrating superior performance in object classification and segmentation compared to state-of-the-art methods. Our work paves the way for further exploration of edge-awareness in 3D point cloud analysis, advancing the frontiers of computer vision research.

\bibliography{ref}

\begin{thebibliography}{39}
\providecommand{\natexlab}[1]{#1}
\providecommand{\url}[1]{\texttt{#1}}
\expandafter\ifx\csname urlstyle\endcsname\relax
  \providecommand{\doi}[1]{doi: #1}\else
  \providecommand{\doi}{doi: \begingroup \urlstyle{rm}\Url}\fi

\bibitem[Wu et~al.(2015)Wu, Song, Khosla, Yu, Zhang, Tang, and Xiao]{wu20153d}
Zhirong Wu, Shuran Song, Aditya Khosla, Fisher Yu, Linguang Zhang, Xiaoou Tang, and Jianxiong Xiao.
\newblock 3d shapenets: A deep representation for volumetric shapes.
\newblock In \emph{Proceedings of the IEEE conference on computer vision and pattern recognition}, pages 1912--1920, 2015.

\bibitem[Chang et~al.(2015)Chang, Funkhouser, Guibas, Hanrahan, Huang, Li, Savarese, Savva, Song, Su, et~al.]{chang2015shapenet}
Angel~X Chang, Thomas Funkhouser, Leonidas Guibas, Pat Hanrahan, Qixing Huang, Zimo Li, Silvio Savarese, Manolis Savva, Shuran Song, Hao Su, et~al.
\newblock Shapenet: An information-rich 3d model repository.
\newblock \emph{arXiv preprint arXiv:1512.03012}, 2015.

\bibitem[Dai et~al.(2017)Dai, Chang, Savva, Halber, Funkhouser, and Nie{\ss}ner]{dai2017scannet}
Angela Dai, Angel~X Chang, Manolis Savva, Maciej Halber, Thomas Funkhouser, and Matthias Nie{\ss}ner.
\newblock Scannet: Richly-annotated 3d reconstructions of indoor scenes.
\newblock In \emph{Proceedings of the IEEE conference on computer vision and pattern recognition}, pages 5828--5839, 2017.

\bibitem[Eldar et~al.(1997)Eldar, Lindenbaum, Porat, and Zeevi]{eldar1997farthest}
Yuval Eldar, Michael Lindenbaum, Moshe Porat, and Yehoshua~Y Zeevi.
\newblock The farthest point strategy for progressive image sampling.
\newblock \emph{IEEE Transactions on Image Processing}, 6\penalty0 (9):\penalty0 1305--1315, 1997.

\bibitem[Dovrat et~al.(2019)Dovrat, Lang, and Avidan]{dovrat2019learning}
Oren Dovrat, Itai Lang, and Shai Avidan.
\newblock Learning to sample.
\newblock In \emph{Proceedings of the IEEE/CVF Conference on Computer Vision and Pattern Recognition}, pages 2760--2769, 2019.

\bibitem[Lang et~al.(2020)Lang, Manor, and Avidan]{lang2020samplenet}
Itai Lang, Asaf Manor, and Shai Avidan.
\newblock Samplenet: Differentiable point cloud sampling.
\newblock In \emph{Proceedings of the IEEE/CVF Conference on Computer Vision and Pattern Recognition}, pages 7578--7588, 2020.

\bibitem[Lin et~al.(2021)Lin, Huang, Zhou, Jiang, Wang, and Lei]{lin2021net}
Yanan Lin, Yan Huang, Shihao Zhou, Mengxi Jiang, Tianlong Wang, and Yunqi Lei.
\newblock Da-net: Density-adaptive downsampling network for point cloud classification via end-to-end learning.
\newblock In \emph{2021 4th International Conference on Pattern Recognition and Artificial Intelligence (PRAI)}, pages 13--18. IEEE, 2021.

\bibitem[Zhang et~al.(2020)Zhang, Li, Song, Xie, and Zhang]{zhang2020fact}
Yicheng Zhang, Lei Li, Li~Song, Rong Xie, and Wenjun Zhang.
\newblock Fact: fused attention for clothing transfer with generative adversarial networks.
\newblock In \emph{Proceedings of the AAAI Conference on Artificial Intelligence}, volume~34, pages 12894--12901, 2020.

\bibitem[Wu et~al.(2019{\natexlab{a}})Wu, Li, and Li]{wu2019fase}
Meng Wu, Lei Li, and Hongyan Li.
\newblock Fase: Feature-based similarity search on ecg data.
\newblock In \emph{2019 IEEE International Conference on Big Knowledge (ICBK)}, pages 273--280. IEEE, 2019{\natexlab{a}}.

\bibitem[Li et~al.(2022{\natexlab{a}})Li, Zhang, Oehmcke, Gieseke, and Igel]{li2022mask}
Lei Li, Tianfang Zhang, Stefan Oehmcke, Fabian Gieseke, and Christian Igel.
\newblock Mask-fpan: Semi-supervised face parsing in the wild with de-occlusion and uv gan.
\newblock \emph{arXiv preprint arXiv:2212.09098}, 2022{\natexlab{a}}.

\bibitem[Thomas et~al.(2019)Thomas, Qi, Deschaud, Marcotegui, Goulette, and Guibas]{thomas2019kpconv}
Hugues Thomas, Charles~R Qi, Jean-Emmanuel Deschaud, Beatriz Marcotegui, Fran{\c{c}}ois Goulette, and Leonidas~J Guibas.
\newblock Kpconv: Flexible and deformable convolution for point clouds.
\newblock In \emph{Proceedings of the IEEE/CVF international conference on computer vision}, pages 6411--6420, 2019.

\bibitem[Zhang et~al.(2023{\natexlab{a}})Zhang, Li, Cao, Pu, and Peng]{zhang2023attention}
Tianfang Zhang, Lei Li, Siying Cao, Tian Pu, and Zhenming Peng.
\newblock Attention-guided pyramid context networks for detecting infrared small target under complex background.
\newblock \emph{IEEE Transactions on Aerospace and Electronic Systems}, 2023{\natexlab{a}}.

\bibitem[Qi et~al.(2017{\natexlab{a}})Qi, Su, Mo, and Guibas]{qi2017pointnet}
Charles~R Qi, Hao Su, Kaichun Mo, and Leonidas~J Guibas.
\newblock Pointnet: Deep learning on point sets for 3d classification and segmentation.
\newblock In \emph{Proceedings of the IEEE conference on computer vision and pattern recognition}, pages 652--660, 2017{\natexlab{a}}.

\bibitem[Qi et~al.(2017{\natexlab{b}})Qi, Yi, Su, and Guibas]{qi2017pointnet++}
Charles~Ruizhongtai Qi, Li~Yi, Hao Su, and Leonidas~J Guibas.
\newblock Pointnet++: Deep hierarchical feature learning on point sets in a metric space.
\newblock \emph{Advances in neural information processing systems}, 30, 2017{\natexlab{b}}.

\bibitem[Yan et~al.(2020)Yan, Zheng, Li, Wang, and Cui]{yan2020pointasnl}
Xu~Yan, Chaoda Zheng, Zhen Li, Sheng Wang, and Shuguang Cui.
\newblock Pointasnl: Robust point clouds processing using nonlocal neural networks with adaptive sampling.
\newblock In \emph{Proceedings of the IEEE/CVF conference on computer vision and pattern recognition}, pages 5589--5598, 2020.

\bibitem[Muzahid et~al.(2020)Muzahid, Wan, Sohel, Wu, and Hou]{muzahid2020curvenet}
AAM Muzahid, Wanggen Wan, Ferdous Sohel, Lianyao Wu, and Li~Hou.
\newblock Curvenet: Curvature-based multitask learning deep networks for 3d object recognition.
\newblock \emph{IEEE/CAA Journal of Automatica Sinica}, 8\penalty0 (6):\penalty0 1177--1187, 2020.

\bibitem[Wang et~al.(2019)Wang, Sun, Liu, Sarma, Bronstein, and Solomon]{wang2019dynamic}
Yue Wang, Yongbin Sun, Ziwei Liu, Sanjay~E Sarma, Michael~M Bronstein, and Justin~M Solomon.
\newblock Dynamic graph cnn for learning on point clouds.
\newblock \emph{Acm Transactions On Graphics (tog)}, 38\penalty0 (5):\penalty0 1--12, 2019.

\bibitem[Xu et~al.(2018)Xu, Fan, Xu, Zeng, and Qiao]{xu2018spidercnn}
Yifan Xu, Tianqi Fan, Mingye Xu, Long Zeng, and Yu~Qiao.
\newblock Spidercnn: Deep learning on point sets with parameterized convolutional filters.
\newblock In \emph{Proceedings of the European conference on computer vision (ECCV)}, pages 87--102, 2018.

\bibitem[Engel et~al.(2021)Engel, Belagiannis, and Dietmayer]{engel2021point}
Nico Engel, Vasileios Belagiannis, and Klaus Dietmayer.
\newblock Point transformer.
\newblock \emph{IEEE Access}, 9:\penalty0 134826--134840, 2021.

\bibitem[Oehmcke et~al.(2022{\natexlab{a}})Oehmcke, Li, Revenga, Nord-Larsen, Trepekli, Gieseke, and Igel]{oehmcke:22}
Stefan Oehmcke, Lei Li, Jaime Revenga, Thomas Nord-Larsen, Katerina Trepekli, Fabian Gieseke, and Christian Igel.
\newblock Deep learning based {3D} point cloud regression for estimating forest biomass.
\newblock In \emph{International Conference on Advances in Geographic Information Systems (SIGSPATIAL)}. {ACM}, 2022{\natexlab{a}}.

\bibitem[Revenga et~al.(2022)Revenga, Trepekli, Oehmcke, Jensen, Li, Igel, Gieseke, and Friborg]{revenga2022above}
Jaime~C Revenga, Katerina Trepekli, Stefan Oehmcke, Rasmus Jensen, Lei Li, Christian Igel, Fabian~Cristian Gieseke, and Thomas Friborg.
\newblock Above-ground biomass prediction for croplands at a sub-meter resolution using uav--lidar and machine learning methods.
\newblock \emph{Remote Sensing}, 14\penalty0 (16):\penalty0 3912, 2022.

\bibitem[Oehmcke et~al.(2022{\natexlab{b}})Oehmcke, Li, Revenga, Nord-Larsen, Trepekli, Gieseke, and Igel]{oehmcke2022deep}
Stefan Oehmcke, Lei Li, Jaime~C Revenga, Thomas Nord-Larsen, Katerina Trepekli, Fabian Gieseke, and Christian Igel.
\newblock Deep learning based 3d point cloud regression for estimating forest biomass.
\newblock In \emph{Proceedings of the 30th International Conference on Advances in Geographic Information Systems}, pages 1--4, 2022{\natexlab{b}}.

\bibitem[Wang et~al.(2021)Wang, Jin, Cen, Lang, and Li]{wang2021pst}
Xu~Wang, Yi~Jin, Yigang Cen, Congyan Lang, and Yidong Li.
\newblock Pst-net: point cloud sampling via point-based transformer.
\newblock In \emph{Image and Graphics: 11th International Conference, ICIG 2021, Haikou, China, August 6--8, 2021, Proceedings, Part III 11}, pages 57--69. Springer, 2021.

\bibitem[Wang et~al.(2022)Wang, Jin, Cen, Wang, Tang, and Li]{wang2022lightn}
Xu~Wang, Yi~Jin, Yigang Cen, Tao Wang, Bowen Tang, and Yidong Li.
\newblock Lightn: Light-weight transformer network for performance-overhead tradeoff in point cloud downsampling.
\newblock \emph{arXiv preprint arXiv:2202.06263}, 2022.

\bibitem[Zhang et~al.(2022)Zhang, Li, Igel, Oehmcke, Gieseke, and Peng]{zhang2022lr}
Tianfang Zhang, Lei Li, Christian Igel, Stefan Oehmcke, Fabian Gieseke, and Zhenming Peng.
\newblock Lr-csnet: Low-rank deep unfolding network for image compressive sensing.
\newblock \emph{arXiv preprint arXiv:2212.09088}, 2022.

\bibitem[Zhang et~al.(2023{\natexlab{b}})Zhang, Li, Cao, Pu, and Peng]{10024907}
Tianfang Zhang, Lei Li, Siying Cao, Tian Pu, and Zhenming Peng.
\newblock Attention-guided pyramid context networks for detecting infrared small target under complex background.
\newblock \emph{IEEE Transactions on Aerospace and Electronic Systems}, pages 1--13, 2023{\natexlab{b}}.
\newblock \doi{10.1109/TAES.2023.3238703}.

\bibitem[Zhou et~al.()Zhou, Yuan, Wang, Li, Oehmcke, Liu, and Peng]{zhou4425635multi}
Changsheng Zhou, Chao Yuan, Hongxin Wang, Lei Li, Stefan Oehmcke, Junmin Liu, and Jigen Peng.
\newblock Multi-scale pseudo labeling for unsupervised deep edge detection.
\newblock \emph{Available at SSRN 4425635}.

\bibitem[Li et~al.(2022{\natexlab{b}})Li, Zhang, Oehmcke, Gieseke, and Igel]{li2022buildseg}
Lei Li, Tianfang Zhang, Stefan Oehmcke, Fabian Gieseke, and Christian Igel.
\newblock Buildseg buildseg: A general framework for the segmentation of buildings.
\newblock \emph{Nordic Machine Intelligence}, 2\penalty0 (3), 2022{\natexlab{b}}.

\bibitem[Wu et~al.(2023)Wu, Zheng, Pfrommer, and Beyerer]{wu2023attention}
Chengzhi Wu, Junwei Zheng, Julius Pfrommer, and J{\"u}rgen Beyerer.
\newblock Attention-based point cloud edge sampling.
\newblock In \emph{Proceedings of the IEEE/CVF Conference on Computer Vision and Pattern Recognition}, pages 5333--5343, 2023.

\bibitem[Li et~al.(2018)Li, Bu, Sun, Wu, Di, and Chen]{li2018pointcnn}
Yangyan Li, Rui Bu, Mingchao Sun, Wei Wu, Xinhan Di, and Baoquan Chen.
\newblock Pointcnn: Convolution on x-transformed points.
\newblock \emph{Advances in neural information processing systems}, 31, 2018.

\bibitem[Wu et~al.(2019{\natexlab{b}})Wu, Qi, and Fuxin]{wu2019pointconv}
Wenxuan Wu, Zhongang Qi, and Li~Fuxin.
\newblock Pointconv: Deep convolutional networks on 3d point clouds.
\newblock In \emph{Proceedings of the IEEE/CVF Conference on computer vision and pattern recognition}, pages 9621--9630, 2019{\natexlab{b}}.

\bibitem[Liu et~al.(2019)Liu, Tang, Lin, and Han]{liu2019point}
Zhijian Liu, Haotian Tang, Yujun Lin, and Song Han.
\newblock Point-voxel cnn for efficient 3d deep learning.
\newblock \emph{Advances in Neural Information Processing Systems}, 32, 2019.

\bibitem[Zhao et~al.(2021)Zhao, Jiang, Jia, Torr, and Koltun]{zhao2021point}
Hengshuang Zhao, Li~Jiang, Jiaya Jia, Philip~HS Torr, and Vladlen Koltun.
\newblock Point transformer.
\newblock In \emph{Proceedings of the IEEE/CVF international conference on computer vision}, pages 16259--16268, 2021.

\bibitem[Cheng et~al.(2021)Cheng, Chen, He, Liu, and Bai]{cheng2021net}
Silin Cheng, Xiwu Chen, Xinwei He, Zhe Liu, and Xiang Bai.
\newblock Pra-net: Point relation-aware network for 3d point cloud analysis.
\newblock \emph{IEEE Transactions on Image Processing}, 30:\penalty0 4436--4448, 2021.

\bibitem[Xu et~al.(2021)Xu, Ding, Zhao, and Qi]{xu2021paconv}
Mutian Xu, Runyu Ding, Hengshuang Zhao, and Xiaojuan Qi.
\newblock Paconv: Position adaptive convolution with dynamic kernel assembling on point clouds.
\newblock In \emph{Proceedings of the IEEE/CVF Conference on Computer Vision and Pattern Recognition}, pages 3173--3182, 2021.

\bibitem[Wiersma et~al.(2021)Wiersma, Nasikun, Eisemann, and Hildebrandt]{wiersma2021deltaconv}
Ruben Wiersma, Ahmad Nasikun, Elmar Eisemann, and Klaus Hildebrandt.
\newblock Deltaconv: Anisotropic point cloud learning with exterior calculus.
\newblock \emph{arXiv preprint arXiv:2111.08799}, 2021.

\bibitem[Su et~al.(2018)Su, Jampani, Sun, Maji, Kalogerakis, Yang, and Kautz]{su2018splatnet}
Hang Su, Varun Jampani, Deqing Sun, Subhransu Maji, Evangelos Kalogerakis, Ming-Hsuan Yang, and Jan Kautz.
\newblock Splatnet: Sparse lattice networks for point cloud processing.
\newblock In \emph{Proceedings of the IEEE conference on computer vision and pattern recognition}, pages 2530--2539, 2018.

\bibitem[Lai et~al.(2022)Lai, Liu, Jiang, Wang, Zhao, Liu, Qi, and Jia]{lai2022stratified}
Xin Lai, Jianhui Liu, Li~Jiang, Liwei Wang, Hengshuang Zhao, Shu Liu, Xiaojuan Qi, and Jiaya Jia.
\newblock Stratified transformer for 3d point cloud segmentation.
\newblock In \emph{Proceedings of the IEEE/CVF Conference on Computer Vision and Pattern Recognition}, pages 8500--8509, 2022.

\bibitem[Qian et~al.(2020)Qian, Hou, Zhang, Zeng, Kwong, and He]{qian2020mops}
Yue Qian, Junhui Hou, Qijian Zhang, Yiming Zeng, Sam Kwong, and Ying He.
\newblock Mops-net: A matrix optimization-driven network fortask-oriented 3d point cloud downsampling.
\newblock \emph{arXiv preprint arXiv:2005.00383}, 2020.

\end{thebibliography}
\end{sloppypar}

\end{document}